\documentclass[sigconf]{acmart}

\AtBeginDocument{%
  \providecommand\BibTeX{{%
    \normalfont B\kern-0.5em{\scshape i\kern-0.25em b}\kern-0.8em\TeX}}}

\copyrightyear{2021}
\acmYear{2021}
\setcopyright{acmcopyright}\acmConference[ICAIF'21]{2nd ACM International Conference on AI in Finance}{November 3--5, 2021}{Virtual Event, USA}
\acmBooktitle{2nd ACM International Conference on AI in Finance (ICAIF'21), November 3--5, 2021, Virtual Event, USA}
\acmPrice{15.00}
\acmDOI{10.1145/3490354.3494404}
\acmISBN{978-1-4503-9148-1/21/11}

\begin{document}

\title{Deep Video Prediction for Time Series Forecasting}

\author{Zhen Zeng}
\affiliation{%
  \institution{J.~P.~Morgan AI Research}
  \city{New York}
  \state{NY}
  \country{USA}}
\email{zhen.zeng@jpmorgan.com}

\author{Tucker Balch}
\affiliation{%
  \institution{J.~P.~Morgan AI Research}
  \city{New York}
  \state{NY}
  \country{USA}}
\email{tucker.balch@jpmorgan.com}

\author{Manuela Veloso}
\affiliation{%
  \institution{J.~P.~Morgan AI Research}
  \city{New York}
  \state{NY}
  \country{USA}}
\email{manuela.veloso@jpmorgan.com}

\renewcommand{\shortauthors}{Zeng Z., et al.}



\begin{abstract}
Time series forecasting is essential for decision making in many domains. In this work, we address the challenge of predicting prices evolution among multiple potentially interacting financial assets. A solution to this problem has obvious importance for governments, banks, and investors. Statistical methods such as Auto Regressive Integrated Moving Average (ARIMA) are widely applied to these problems. In this paper, we propose to approach economic time series forecasting of multiple financial assets in a novel way via video prediction. Given past prices of multiple potentially interacting financial assets, we aim to predict the prices evolution in the future. Instead of treating the snapshot of prices at each time point as a vector, we spatially layout these prices in 2D as an image similar to market change visualization, and we can harness the power of CNNs in learning a latent representation for these financial assets. Thus, the history of these prices becomes a sequence of images, and our goal becomes predicting future images. We build on advances from computer vision for video prediction. Our experiments involve the prediction task of the price evolution of nine financial assets traded in U.S. stock markets. The proposed method outperforms baselines including ARIMA, Prophet and variations of the proposed method, demonstrating the benefits of harnessing the power of CNNs in the problem of economic time series forecasting.
\end{abstract}

\maketitle

\begin{figure}[t]
\centering
\includegraphics[width=0.47\textwidth]{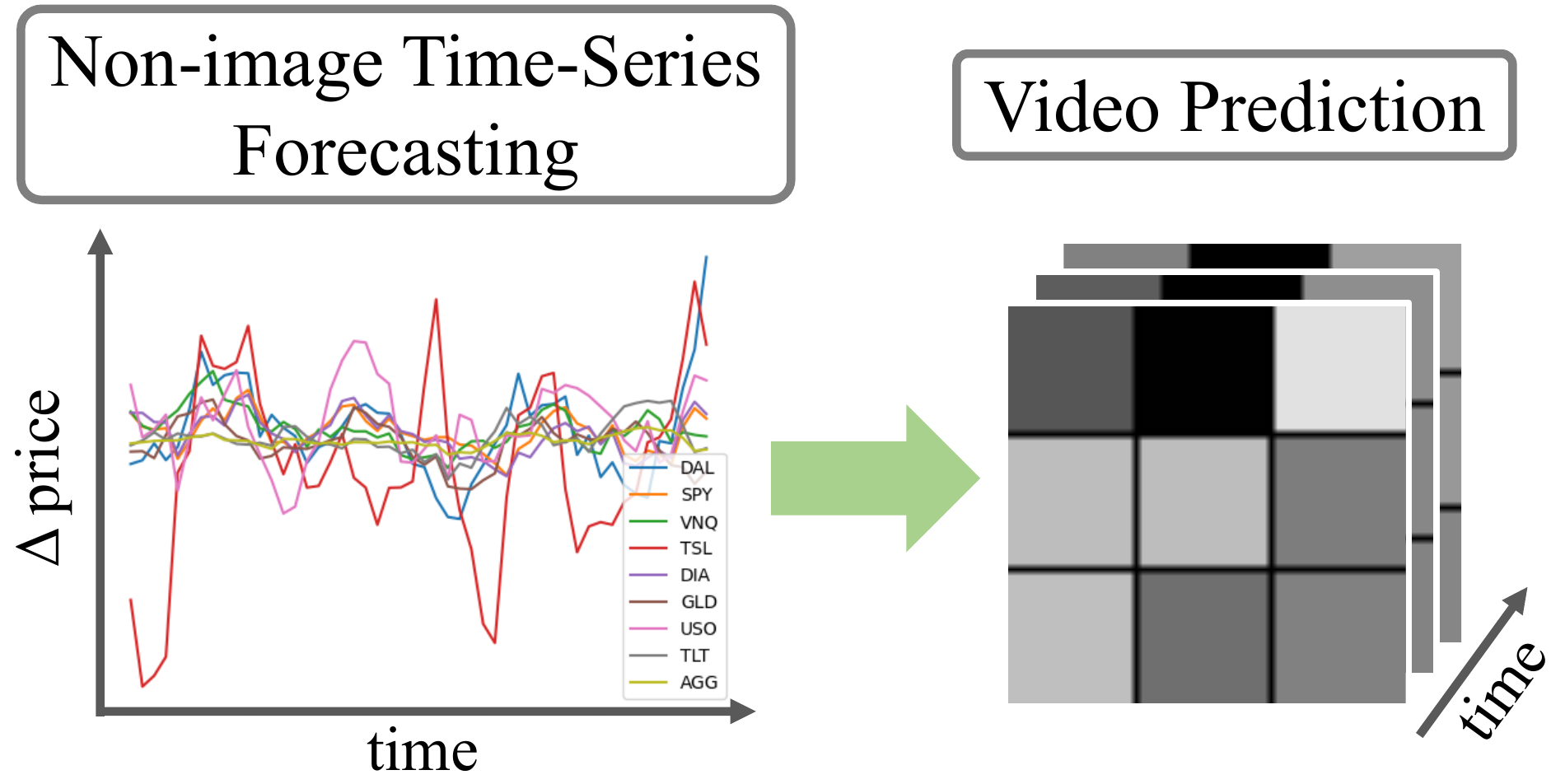}
\caption{We transform the problem of non-image time series forecasting into the problem of video prediction. For example, given a task of forecasting the relative close price change of 9 different publicly traded assets in the future, our approach first transforms the numerical data at each time stamp into an image frame (e.g., a 3x3 tile heatmap), and predicts future image frames via a video prediction technique.}
\label{fig:teaser}
\end{figure}

\section{Introduction}
In many time series forecasting tasks such as predictions on market, sales, and weather, the underlying data is non-image. Common statistical methods such as ARIMA have been widely adopted across these domains for time series forecasting tasks. These methods consider the history of the numerical data and predict the future values of the observed data. On the other hand, given tables or lists of numerical data, humans rely much more on visualizing the underlying numerical data rather than directly eyeing at the numbers themselves to develop a high-level understanding of the data. For example, experienced traders develop intuition for making buy/sell decisions by observing visual market charts~\cite{cohen2019trading}.

The power of visualizations lies in that they provide spatial structural information~\cite{sharma2019deepinsight} when laying out the underlying data in 2D images, which is not available in the original data. When looking at 2D images, human eyes are proficient at capturing spatial structure or patterns to help make better decisions or predictions. Advances in deep learning and computer vision have shown that Convolutional Neural Networks (CNNs)~\cite{NIPS2012_4824} carry the capabilities to extract features of local spatial regions, which enables systems to recognize spatial patterns such as those in object detection and recognition tasks.

\begin{figure*}[t!]
\centering
\includegraphics[width=1.0\textwidth]{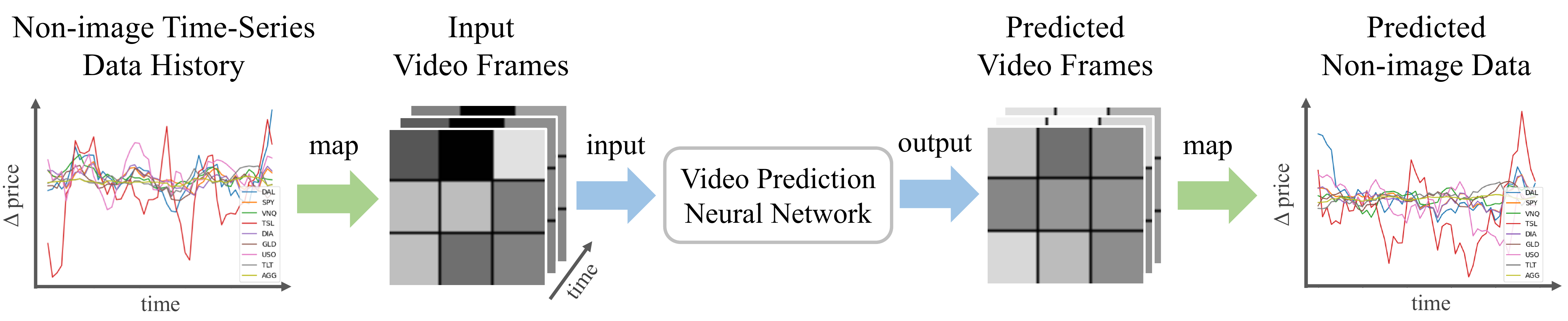}
\caption{Method overview. First, we turn non-image time-series data history into a video frame at each time stamp. Then, we use a video prediction neural network to predict future video frames. Finally, we map the predicted video frames back to the numerical data space.}
\label{fig:overview}
\end{figure*}

Inspired by how humans benefit from 2D visualizations of numerical data, we propose to spatially layout numerical information in 2D images similarly to market change visualizations. Then we take the advantage of CNNs for time series forecasting tasks, which were originally studied in non-image domains. In this paper, we take a unique perspective in predicting non-image time-series data through the lens of computer vision. To achieve this, we propose to first visualize the multivariate time-series data as a sequence of images, thus forming a video, and then build on video prediction techniques~\cite{franceschi2020stochastic,babaeizadeh2017stochastic} to predict future image frames, i.e., future visualizations of the underlying non-image data.

Our experiments focus on the task of forecasting market changes over time, where 9 publicly traded assets including 2 stocks and 7 Exchange-Traded Funds (ETFs) are being considered as shown in Figure \ref{fig:teaser}. We demonstrate that our proposed method outperforms other baselines, such as DeepInsight~\cite{sharma2019deepinsight}, ARIMA and Prophet (on non-image numerical data), as well as variations of our proposed method. Our study shows that our method is able to learn high-level knowledge jointly over multiple assets, and produces better prediction accuracy compared to either learning each asset independently, or learning multiple assets as a vector other than a 2D image.

\section{Related Work}\label{sec:related_work}
Time series forecasting~\cite{hyndman2018forecasting} has many applications across diverse domains, e.g. finance, climate, resource allocation, etc. Among a collection of statistical tools, \textbf{exponential smoothing} and \textbf{ARIMA} are two of the most widely adopted approaches for time series forecasting. Exponential smoothing predicts the future value of a random variable by a weighted average of its past values, where the weight associated with each past value decreases exponentially as we go back in time. Several variations of exponential smoothing are proposed to consider trend~\cite{gardner1985forecasting} and seasonality~\cite{holt2004forecasting,winters1960forecasting} in the data. ARIMA combines autoregressive and moving average models for forecasting and ARIMAs use differencing to help reduce trend and seasonality in the data. However, ARIMA, as well as VAR (vector autoregressive) model for multivariate cases, cannot capture nonlinear patterns in time series, rendering it insufficient for forecasting when nonlinear patterns occur. Our experiments show that a neural network based approach, which is nonlinear, outperforms ARIMA.
    

Recent works~\cite{zhang2003time,pai2005hybrid,safari2018oil} focus on combinations of statistical and machine learning methods to improve forecasting accuracy. Yet the data involved in these time series forecasting tasks~\cite{makridakis2018m4} is usually non-image.

In this work, we provide a new perspective of the time series forecasting problem, by transforming it into a video prediction problem. We visualize the underlying numerical data as an image at each time stamp, and bring recent advances in video prediction~\cite{oprea2020review} from the field of computer vision for forecasting.
    
Early works on video prediction directly predict future appearance as a composition of predicted image patches~\cite{ranzato2014video}, without explicit modeling of temporal dynamics in the video. Others have attempted to learn explicit transformations (ae.g. per pixel motion~\cite{reda2018sdc}, or affine transformations~\cite{finn2016unsupervised}) that synthesize the future frame from the last observed frame. These works learn to infer the transformation parameters from observed video frames. More recently, researchers aim to disentangle motion and visual content~\cite{hsieh2018learning,franceschi2020stochastic} in videos. The prediction task becomes more tractable because prediction can be performed in a latent space modeling the temporal dynamics. Thus, we built on~\cite{franceschi2020stochastic} for video prediction. 
    

Similar to our work, there are recent works that also tackle classification or regression tasks on non-image data from the computer vision perspective. \cite{cohen2019trading,du2020image} developed image classifiers for stock analysis, e.g. buy/sell, and positive/negative price change. \cite{sharma2019deepinsight} proposed DeepInsight which visualizes non-image data as an image through dimension reduction technique, and trained CNNs for classification on cancer types given visualized gene expressions. Although DeepInsight achieves promising classification accuracy as discussed in~\cite{sharma2019deepinsight}, our experiment suggests that DeepInsight is not necessarily suitable for prediction tasks. \cite{li2020forecasting} proposed to transform non-image data into recurrence images, and then use CNNs to extract image features to predict weights for averaging multiple statistical forecasting methods. However, their method is concerned only with univariate time series forecasting. Real-world problems often involve multivariate time series forecasting, and it's important to understand the relationships between multiple variables. Our work addresses multivariate time series forecasting as discussed in the experiments.


\begin{figure*}[t]
\centering
\includegraphics[width=0.6\textwidth]{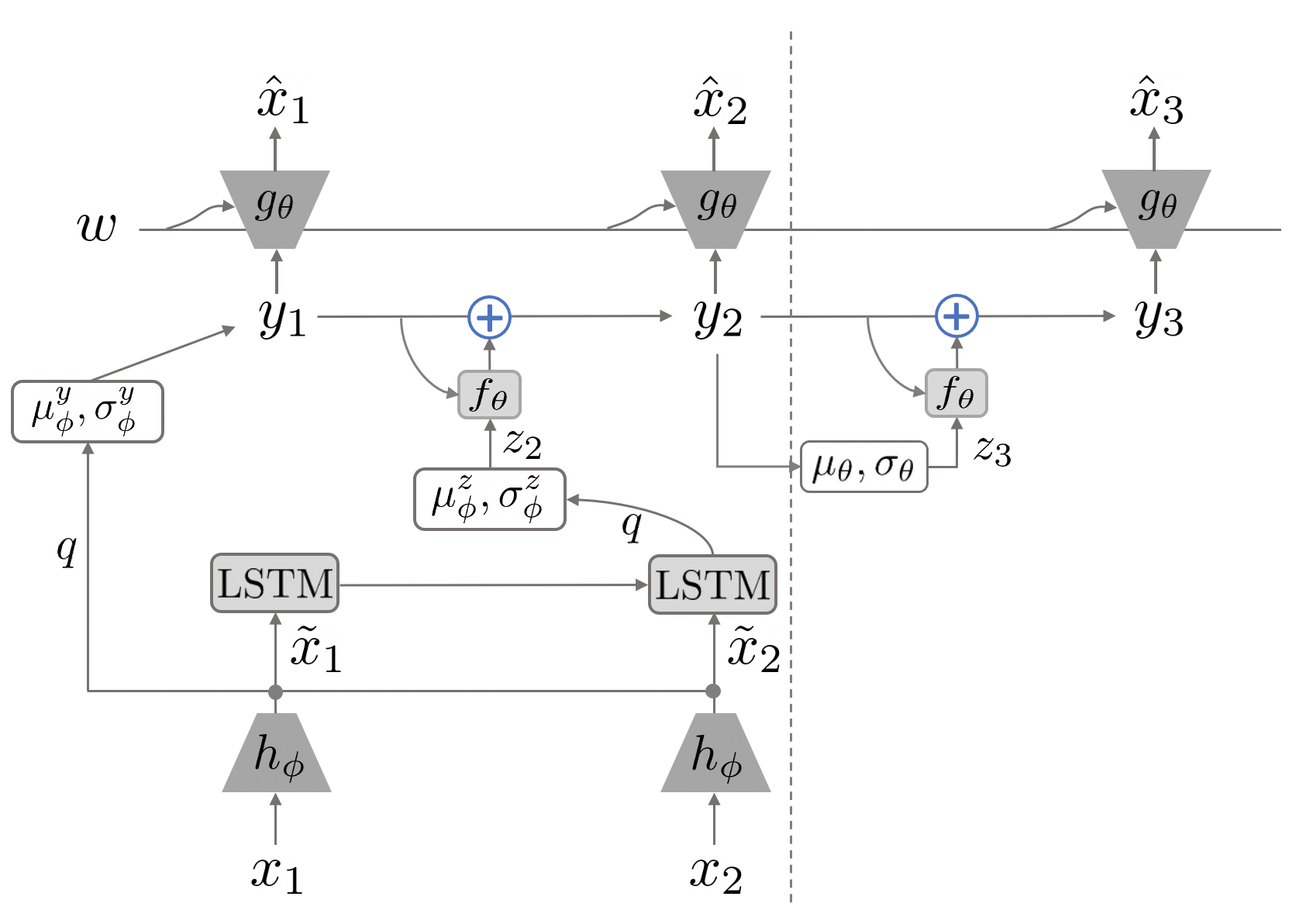}
\caption{Video prediction network. Variational autoencoders are learned for encoding and decoding the input and output video frames. LSTM learns the temporal features of underlying dynamics. Figure recreated from ~\cite{franceschi2020stochastic}.}
\label{fig:srvp}
\end{figure*}

\section{Methods}\label{sec:methods}

Given a time series of a random variable $\{\mathbf{x}_0, \mathbf{x}_1, \cdots, \mathbf{x}_t\}$, where $\textbf{x}_t \in \mathbb{R}^d$, the goal is to predict the values of the random variable at future time stamps $\{\mathbf{x}_{t+1}, \mathbf{x}_{t+2}. \cdots\}$. In this work, for each time $t$, we visualize $\mathbf{x}_t$ as an image, i.e., $\mathbf{x}_t \mapsto \mathcal{I}_t$. Then the task of predicting $\mathbf{x}_{t+\Delta t}$ is converted to a video prediction task, i.e., predicting future image frames $\mathcal{I}_{t+\Delta t}$ given an input video clip. We will explain how we spatially layout data in 2D and the video prediction method as following.

\subsection{Visualization}\label{sec:visualization}
The particular visualizations of the underlying data varies across domains. Humans usually rely on domain knowledge to develop visualizations of numerical data, and these visualizations evolve over time as humans continuously improve them. We do not aim to provide an unified way to visualize all numerical data across arbitrary domains. We believe domain knowledge is important (as we will also show in experiments), and aiming to provide an unified way of visualizations poses the risk of throwing away domain knowledge.

Regardless of domains, the rule of thumb for generating visualizations of time-series data is that correlated data shall be visualized in a way such that they are spatially close to each other in the 2D image. The intuitive reason for this principle is that when looking at visualizations, CNN is good at extracting structural features in local regions represented by its limited receptive field. By visualizing correlated data (i.e. potentially dependent) spatially close in an image, CNN gets the chance to learn the high-level joint features of these correlated data, which can be exploited for prediction tasks. As we later show in our experiments, separating correlated data in visualizations results in a drop of prediction accuracy.

Here we discuss the particular visualization that we use for the market change prediction task in our experiments. Given the time-series market history of 9 assets in terms of relative percentage change $\delta$ of close values, i.e., $\mathbf{x}_t=[\delta_t^1, \cdots, \delta_t^9]^T$, the goal is to predict future percentage changes of these assets. Following the same intuition of a commonly adopted market change visualization in industry from Finviz\footnote{\url{https://finviz.com/map.ashx?t=sec_all}}, we visualize the percentage changes of these 9 assets in a 3x3 tile heatmap, as shown in Figure \ref{fig:teaser}. We will discuss how domain knowledge helps arrange these 9 assets and achieve better performance in experiments and discussion. To visualize numerical data into pixels, we convert the percentage change of $i$th asset at time $t$, $\delta_t^i$, into a pixel $p \in \mathcal{I}_t$,
\begin{align}
p &= S(\delta_t^i)*255 \label{eq:map} \\
S(x) &= \frac{1}{1+e^{-x}}
\end{align}
where $S: \mathcal{R} \mapsto [0, 1]$ is a sigmoid function, thus $p \in [0, 255]$. For example, if $\delta_t^i=3$, meaning that $i$th asset has a $3\%$ increase in its close value at time $t$, then the corresponding pixel value will be 243. As such, the higher the percentage increase, the brighter the visualized pixel. And the more the percentage decrease, the darker the visualized pixel.

\begin{figure*}[t!]
\centering
\includegraphics[width=1.0\textwidth,height = 0.265\textwidth]{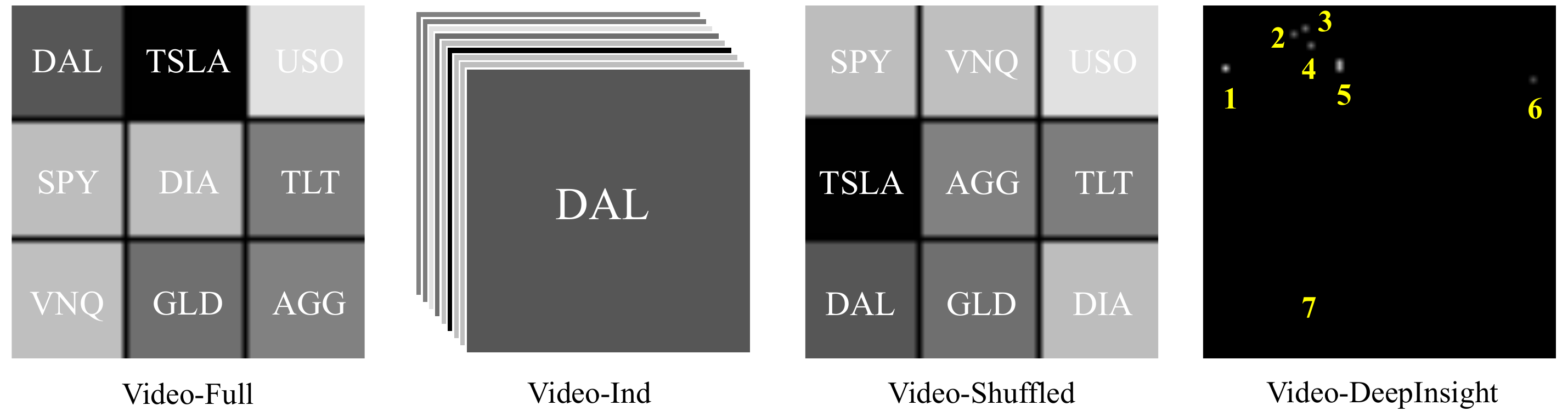}
\caption{Visualization examples used in different methods. \textbf{Video-Full}: correlated assets are placed close to each other in the same image; \textbf{Video-Ind}: each asset is visualized independently in separate images; \textbf{Video-Shuffled}: correlated assets are placed apart from each other in the same image; \textbf{Video-DeepInsight}: assets visualized using method proposed in~\protect\cite{sharma2019deepinsight}, where each asset corresponds to a single pixel on the image, resulting in sparse points in the image. Specifically, 1 is USO, 2 is GLD, 3 is TLT, 4 is AGG, region 5 includes SPY, DIA, VNQ, 6 is DAL, 7 is TSLA (dark pixel). Note that DIA and VNQ gets placed at the same pixel location based on the proposed method in~\protect\cite{sharma2019deepinsight}. As a result, we can only visualize one of the two assets and overwrite the other one, which prevents the network from learning the overwritten asset (VNQ is overwritten in our experiments).}
\label{fig:visualization}
\end{figure*}

\subsection{Video Prediction}

In this work, we adapted a video prediction network SRVP (Stochastic Latent Residual Video Prediction)~\cite{franceschi2020stochastic} in computer vision for the economic time series forecasting task. Compared to most works in the literature which rely on image-autoregressive recurrent networks, SRVP decouples frame synthesis and video dynamics estimation. SRVP has shown to outperform prior state-of-the-art methods across a simulated~\cite{denton2018stochastic} dataset, and real-world datasets of human activities~\cite{schuldt2004recognizing,ionescu2013human3} as well as robot actions~\cite{ebert2017self}. We adapted the video prediction network from predicting frames in natural video to predicting frames in visualizations.

As shown in Figure~\ref{fig:srvp}, SRVP explicitly models the hidden state $y_t$ as well as the dynamics (the residual gets added to $y_t$ from $f_\theta$ at each time point) of the video in a latent space. Specifically, $x_t$ denotes the input frame at time $t$, $y_t$ is the latent state variable, $z_t$ is the latent state dynamics variable, and $w$ is a content variable that encodes the static content in the video (e.g. static background, constant shape of foreground object, etc).

For each input frame $x_t$, $h_\phi$ is an CNN-based encoder that encodes $x_t$ into encoded frame $\tilde{x}_t$. Given encoded frames $\tilde{x}_{1:k}$ of the input video, the initial latent state $y_1 \sim \mathcal{N}(\mu_\phi^y, \sigma_\phi^y)$ is obtained through variational inference. We use $k=5$ across our experiments. The latent state $y_t$ then propagates forward with a transition function $f_\theta$,
\[ y_{t+1} = y_t + f_\theta (z_{t+1}) \]
where $f_\theta$ is an Multilayer Perception (MLP) that learns the first-order movement of latent state $y_t$. As part of the input to $f_\theta$, the latent dynamics $z_t \sim \mathcal{N}(\mu_\phi^z, \sigma_\phi^z)$ is inferred through an LSTM on the encoded input frames. The content variable $w$ is inferred through a permutation-invariant function~\cite{zaheer2017deep} given few encoded frames. Lastly, $g_\theta$ is a decoder network that concatenates the content variable $w$ and latent state $y_t$ and decode it back to the original space of $x_t$, thus producing the estimated $\hat{x}_t$. Note that at testing time, we need to predict future $z_t$ when $x_t$ is not available. Instead of inferring $z_t$ through LSTM on $x_t$s as explained earlier, $z_t$ is inferred from $y_{t-1}$ through two MLPs to generate $\mu_\theta, \sigma_\theta$, which are trained to fit $z_{t+1} \sim \mathcal{N}(\mu_\theta(y_t), \sigma_\theta(y_t)I)$.

In our experiments, we used VGG16~\cite{simonyan2014very} as the encoder network $h_\phi$ and decoder network $g_\theta$ (mirrored VGG16). The image size is 64x64, and we choose to keep this generic choice so that we don't tailor it towards the specific task in our experiments. We used 50 dimensions for both $y_t$ and $z_t$. The loss function is negative log-likelihood and we used L2 regularization to prevent overfitting. We used Adam for optimization during training, with learning rate $\alpha=3\text{e}^{-3}$, and $\beta_1=0.9, \beta_2=0.999, \epsilon=1\text{e}^{-8}$. 


\section{Experiments}
The experiments were conducted on a Linux machine with Intel(R) Xeon(R) Platinum 8259CL CPU @ 2.50GHz, 19 GB RAM, and NVIDIA T4 GPU. Our experiments focus on the task of forecasting market changes. We consider 9 assets in the market:
\begin{itemize}
    \item DAL (Delta Air Lines, Inc)
    \item SPY (SPDR S\&P 500 ETF Trust)
    \item VNQ (Vanguard Real Estate Index Fund ETF Shares)
    \item TSLA (Tesla, Inc.)
    \item DIA (SPDR Dow Jones Industrial Average ETF Trust)
    \item GLD (SPDR Gold Shares)
    \item USO (United States Oil Fund, LP)
    \item TLT (iShares 20+ Year Treasury Bond ETF)
    \item AGG (iShares Core U.S. Aggregate Bond ETF)
\end{itemize}

We deliberately selected a diverse group of assets. There exists interdependencies between these assets. For instance, the airline stock DAL will usually go up in price when the oil ETF USO goes down in price.  This is because fuel derived from oil is one of the primary operating costs for airlines, and a decrease in fuel prices can be predictive of future profits.  Similarly, SPY which represents large U.S. company stocks is typically inversely related to the movement of TLT which represents long term bonds.  Other assets in the mix share other correlations or anti-correlations that reflect the structure of the U.S. economy. We show that our method is able to learn and exploit these hidden interdependencies to make joint predictions.

In our experiments, we used the closing prices of these assets from June 29th, 2010 to Dec 31rd, 2019 (source: Yahoo!-Finance\footnote{\url{https://finance.yahoo.com/}}). We pre-processed the collected data by calculating the percentage change of each asset's closing price on each day with respect to its closing price 5 days ago. The task is that given the percentage changes of assets over 5 consecutive days, we need to predict the future percentage changes of assets for the next 10 days. We split the historical data from June 29th, 2010 to Dec 31rd, 2018 for training and validation (we did 95\% split for training and validation), and Jan 1st, 2019 to Dec 31rd, 2019 for testing. We benchmarked the prediction performance of our proposed video prediction base method (referred as \textbf{Video-Full} below), against baseline methods including ARIMA and Prophet on original non-image data, as well as variations of the proposed method. Specifically,

\begin{figure}[t!]
\centering
\includegraphics[width=0.48\textwidth]{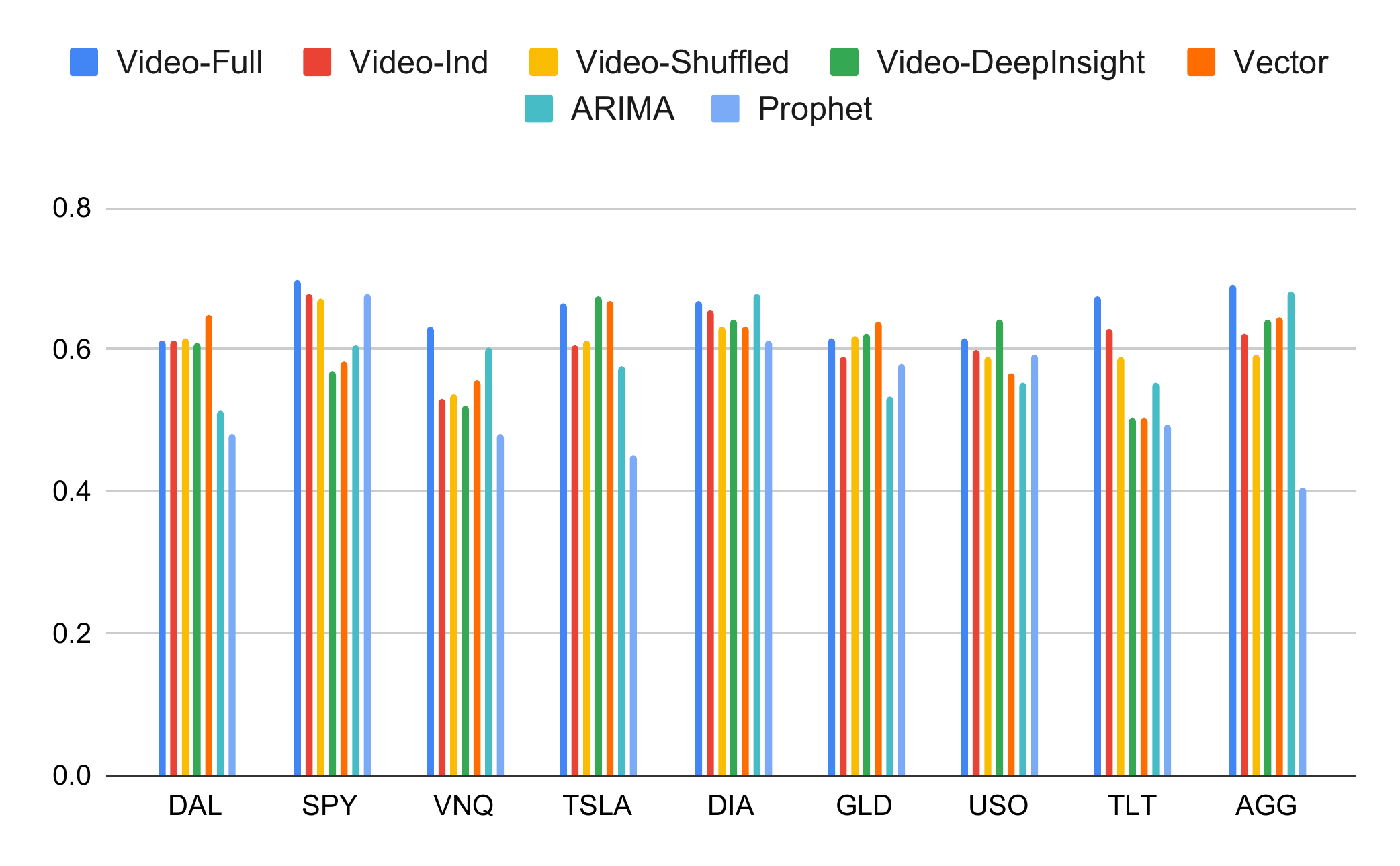}
\caption{Prediction accuracy of proposed method against baselines with $\lambda=0.5$ in evaluation metric (Eq. \ref{eq:metric}).}
\label{fig:lambda05}
\end{figure}

\begin{itemize}
    \item \textbf{Video-Full}: We turn the numerical data of market percentage change into visualizations of 3x3 tile heatmap as explained in section \ref{sec:visualization}. Then 5 frames are given as input to the SRVP neural network, which outputs predicted 10 future frames. Thus, the learned SRVP neural network predicts all 9 assets market change \textbf{jointly}. Note that the 3x3 tile arrangements of the 9 assets are based on domain knowledge, such that known to be correlated assets are placed close to each other.
    \item \textbf{Video-Ind}: Instead of visualizing all 9 assets in a 3x3 tile heatmap and learn to predict jointly as in Video-Full, here we turn each asset market percentage change into a single tile heatmap, thus producing one video clip for each asset. For each asset, we \textbf{independently} train a SRVP neural network to predict its future percentage change.
    \item \textbf{Video-Shuffled}: This baseline method is the same as Video-Full, except that 3x3 tile arrangements of the 9 assets are shuffled such that it goes against the domain knowledge, meaning that known to be correlated assets are placed apart from each other.
    \item \textbf{Video-DeepInsight}: As discussed in section \ref{sec:related_work}, DeepInsight \cite{sharma2019deepinsight} takes a similar perspective of converting non-image data to image data, and adopt computer vision techniques on the image data. Here we use their proposed method to visualize the non-image data, by corresponding each dimension of the non-image data with a 2D pixel coordinate on the image plane through PCA. We then apply SRVP on the resulting visualizations.
    \item \textbf{Vector}: Instead of turning numerical data into 2D heatmaps as in the proposed method, here we directly use the original numerical data (normalized to $[0,1]$) as a 9x1 vector. In order to use the same SRVP network architecture as in other baselines for fair comparison, we repeat the 9x1 vectors to match the same size of input used in other baselines. Note that this method can be perceived as a neural network based nonlinear vector autoregressive model.
    \item \textbf{ARIMA}: For each asset, we directly apply ARIMA to the non-image numerical data to predict its future percentage changes. We used auto ARIMA\footnote{\url{https://alkaline-ml.com/pmdarima/index.html}} to search for a proper set of parameters for each asset and fit the data.
     \item \textbf{Prophet}: For each asset, we use Prophet\footnote{\url{https://facebook.github.io/prophet/}} library from Facebook designated for time series forecasting. We applied Prophet on the original numerical data to predict its future percentage changes.
\end{itemize}

\begin{figure}[!t]
\centering
\includegraphics[width=0.48\textwidth]{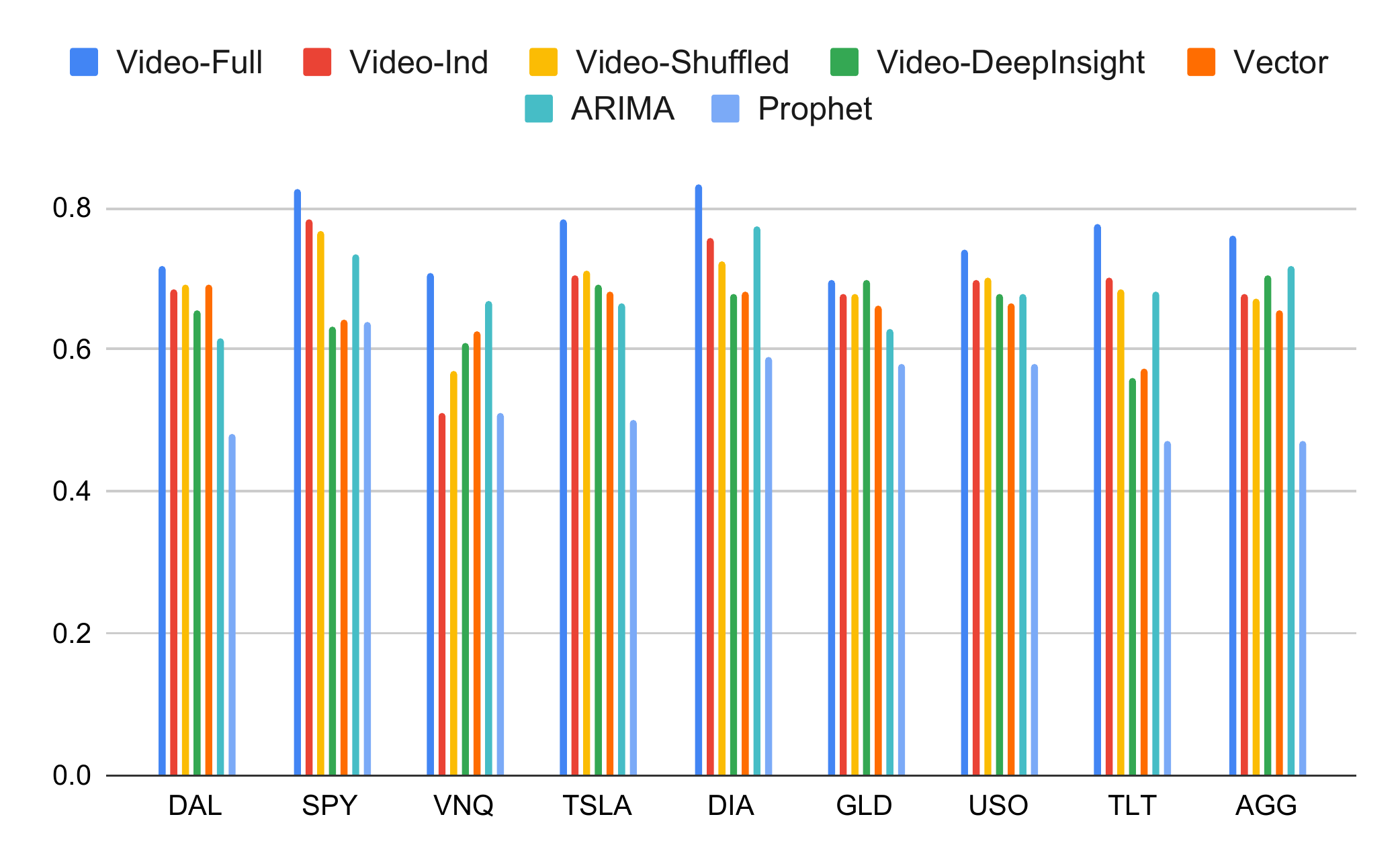}
\caption{Prediction accuracy of proposed method against baselines with $\lambda=10$ in evaluation metric (Eq. \ref{eq:metric}).}
\label{fig:lambda10}
\end{figure}

Examples of visualizations based on these discussed methods are as shown in Figure \ref{fig:visualization}. We present only visualizations for methods that do involve a visualization step and learn in image domain, thus excluding \textbf{Vector}, \textbf{ARIMA} and \textbf{Prophet}.

Our goal of predicting asset price changes is not to automate the process of trading, but to augment practitioners during decision making. When looking at asset price history, practitioners can benefit from predictions of whether the price will go up or down (reflected by the sign of the relative percentage changes). Thus, in order to measure the prediction performance, we evaluate the accuracy of the predicted sign of asset relative percentage change, defined as
\begin{align}
    Acc &= \sum_k \sum_i w_k \mathbb{I}(\delta_k^i, \hat{
    \delta}_k^i) \\
    w_k &= e^{-\lambda k} \label{eq:metric}
\end{align}
where $\mathbb{I}(\delta_k^i, \hat{\delta}_k^i)$ is 1 if $\delta_k^i, \hat{\delta}_k^i$ share the same sign, otherwise 0.

\begin{table*}[!t]
\begin{tabular}{|c|c|c|c|c|c|c|c|}
\hline
                    & \textbf{Video-Full} & Video-Ind & Video-Shuffled & Video-DeepInsight & Vector & ARIMA & Prophet \\
\hline
$\lambda=0.5$  & $\mathbf{0.65\pm0.03}$ & $0.61\pm0.04$ & $0.61\pm0.04$ & $0.60\pm0.06$ & $0.61\pm0.05$ & $0.59\pm0.06$ & $0.53\pm0.09$\\
\hline
$\lambda=10$ & $\mathbf{0.76\pm0.05}$ & $0.69\pm0.08$ & $0.69\pm0.05$ & $0.66\pm0.05$ & $0.65\pm0.04$ & $0.65\pm0.04$ & $0.54\pm0.06$ \\
\hline
\end{tabular}
\caption{Benchmark of prediction accuracy averaged over all assets in experiments.}
\label{tab:table}
\end{table*}

For each asset, we convert the predicted pixel values averaged within the corresponding tile in the heatmap back to numerical data $\hat{\delta}_k^i$ using the inverse of Eq. \ref{eq:map}. Note that we use the exponentially decaying weight $w_k$ to weight the prediction accuracy. $k=0$ indicates the 1st day into the future following the given days in the input. This is to reflect that it is more important to achieve a better prediction in the near future compared to far into the future.

We show the prediction performance for each asset in Figure \ref{fig:lambda05}, \ref{fig:lambda10}, and Table~\ref{tab:table}. In Figure \ref{fig:lambda05}, $\lambda=0.5$ (Equation~\ref{eq:metric}), thus the weights $w_k$ of prediction accuracy at each future time stamp are $1.0, 0.61, 0.37, 0.22$, $0.14, \cdots$, which indicates that the prediction accuracy starting on the 5th day into the future do not matter as much. In Figure~\ref{fig:lambda10}, $\lambda=10$ (Equation~\ref{eq:metric}), thus the weights $w_k$ of prediction accuracy at each future time stamp are $1.0, 0.0, \cdots$, indicating that we focus on the prediction performance for the very next future day. Table~\ref{tab:table} summarizes the prediction accuracy averaged over all 9 assets for each method.


As we can see, for either $\lambda$ value, \textbf{Video-Full} outperforms other baseline methods across all 9 assets. More importantly, we show that when learning to predict the market changes jointly, we achieve better prediction performance in \textbf{Video-Full} compared to \textbf{Video-Ind}, which learns to predict the change of each asset independently. This is because \textbf{Video-Full} allows the network to learn and exploit the joint dynamics of these assets, where the interdependencies between these assets play an important role.

We also show that \textbf{Video-Full} outperforms \textbf{Video-DeepInsight} \cite{sharma2019deepinsight}. DeepInsight was originally proposed for classification tasks, and it corresponds each asset to a single pixel during visualization, resulting in a sparse set of points in the image (as shown in Figure \ref{fig:visualization}). A key issue is that this method can lead to different assets being visualized at the same pixel locations, thus pixel location conflicts, and one has to retain one of the assets information and discard the others at such conflicted pixel locations.
Although~\cite{sharma2019deepinsight} showed \textbf{DeepInsight} to be suitable for classification tasks, we have shown that it is not necessarily suitable for prediction tasks, due to the sparse visualization and especially pixel location conflicts. We will discuss the comparison between \textbf{Video-Full} and \textbf{Video-Shuffled} in detail in the later section \ref{sec:domain_know}.

Without the 2D structural information from visualized images, we can see that \textbf{Vector}, \textbf{ARIMA} and \textbf{Prophet} lead to less prediction accuracy than \textbf{Video-Full} in general. This suggests that by turning non-image time-series forecasting into a video prediction problem, we have introduced informative 2D spatial structure in the visualized images, which can be leveraged by CNNs in \textbf{Video-Full} for forecasting.

\section{Discussion}\label{sec:domain_know}
When comparing \textbf{Video-Shuffled} against \textbf{Video-Full}, we can clearly see a drop in prediction performance in \textbf{Video-Shuffled}. This is because \textbf{Video-Shuffled} suffers from the poor 3x3 tile arrangements of those 9 assets, where correlated assets are placed apart from each other in the visualization. On the contrary, \textbf{Video-Full} uses domain knowledge in finance to guide the 3x3 tile arrangements of those 9 assets. For instance, we place SPY and DIA, which represent the similar S\&P 500 stock index and the Dow Jones Industrial index respectively, next to one another. TLT and AGG, which represent large bond indexes are also placed adjacently.

Taking the advantage of domain knowledge, \textbf{Video-Full} places correlated assets close to each other during visualization, and achieves a better prediction performance because CNNs are able to extract high-level structural feature from local regions. As we scale up, one interesting future direction is to learn to spatially layout multivariate non-image data in 2D, with inductive bias from domain knowledge if available.

Although during the experimented period (2010-2019) the overall market goes up as the long-term trend, for the short-term day-to-day price of each asset, the price is not always going up. In particular, across our test dataset, the percentage of time when price goes up is [0.54, 0.70, 0.62, 0.56, 0.63, 0.61, 0.61, 0.55, 0.62], corresponding respectively to assets DAL, SPY, VNQ, TSLA, DIA, GLD, USO, TLT, AGG. That means, if we take a naive predictor that always predicts the prices to go up for the next day for all assets, then the prediction accuracy will be [0.54, 0.70, 0.62, 0.56, 0.63, 0.61, 0.61, 0.55, 0.62]. As shown in Figure~\ref{fig:lambda10}, the prediction accuracy is [0.72, 0.83, 0.71, 0.78, 0.83, 0.70, 0.74, 0.78, 0.76], which does provide a significant percentage improvement of [0.33, 0.19, 0.15, 0.39, 0.32, 0.15, 0.21, 0.42, 0.22] over the naive predictor for each asset respectively. 

\section{Conclusion}
In this paper, we demonstrate the benefit of learning to predict multivariate non-image time-series data in the 2D image domain. By spatially laying out original non-image data in 2D images, we convert the problem of time series forecasting into a video prediction problem. We then adapt recent state-of-the-art video prediction technique from computer vision to the domain of economic time series forecasting. In our experiments, we show that the proposed method is able to learn spatial structural information from the visualizations and outperforms other baseline methods in predicting future market changes. We provide a proof of concept that, by spatially laying out non-image data in 2D, we can harness the power of CNNs and the proposed method outperforms other methods that either treat each dimension of the multivariate data independently, or treat the multivariate data as a vector. This motivates an interesting future direction of learning to spatially layout non-image data in 2D for multivariate time-series forecasting problems.

\subsubsection*{\textbf{Disclaimer:}}
This paper was prepared for information purposes by the Artificial Intelligence Research group of J.~P.~Morgan Chase \& Co.~and its affiliates (“J.~P.~Morgan”), and is not a product of the Research Department of J.~P.~Morgan. J.~P.~Morgan makes no representation and warranty whatsoever and disclaims all liability, for the completeness, accuracy or reliability of the information contained herein.  This document is not intended as investment research or investment advice, or a recommendation, offer or solicitation for the purchase or sale of any security, financial instrument, financial product or service, or to be used in any way for evaluating the merits of participating in any transaction, and shall not constitute a solicitation under any jurisdiction or to any person, if such solicitation under such jurisdiction or to such person would be unlawful.


\bibliographystyle{ACM-Reference-Format}
\bibliography{main.bib}

\end{document}